\documentclass{article}

\usepackage{microtype}
\usepackage{graphicx}
\usepackage{subcaption}
\usepackage{booktabs} 

\usepackage{hyperref}

\usepackage[preprint]{icml2026}

\usepackage{amsmath}
\usepackage{amssymb}
\usepackage{mathtools}
\usepackage{amsthm}

\usepackage[capitalize,noabbrev]{cleveref}

\theoremstyle{plain}

\theoremstyle{definition}

\theoremstyle{remark}

\usepackage[textsize=tiny]{todonotes}

\usepackage[utf8]{inputenc} 
\usepackage[T1]{fontenc}    
\usepackage{hyperref}       
\usepackage{url}            
\usepackage{booktabs}       
\usepackage{amsfonts}       
\usepackage{nicefrac}       
\usepackage{microtype}      
\usepackage{xcolor}         
\usepackage[english]{babel}
\usepackage{soul} 

\usepackage{amsmath}
\usepackage{graphicx}
\usepackage{pifont}
\usepackage{amssymb}
\usepackage{cleveref}
\usepackage{multirow}
\usepackage{tcolorbox}
\usepackage{graphicx}

\usepackage{makecell}
\usepackage{booktabs}
\icmltitlerunning{TS-Agent: Understanding and Reasoning Over Raw Time Series via Iterative Insight Gathering}

\begin{document}

\twocolumn[
  \icmltitle{TS-Agent: Understanding and Reasoning Over Raw Time Series \\ via Iterative Insight Gathering}



  \icmlsetsymbol{was}{*}

  \begin{icmlauthorlist}
    \icmlauthor{Penghang Liu}{jpmc}
    \icmlauthor{Elizabeth Fons}{jpmc}
    \icmlauthor{Annita Vapsi}{jpmc}
    \icmlauthor{Mohsen Ghassemi}{jpmc}
    \icmlauthor{Svitlana Vyetrenko}{comp,was}
    \icmlauthor{Daniel Borrajo}{jpmc}
    \icmlauthor{Vamsi K. Potluru}{jpmc}
    \icmlauthor{Manuela Veloso}{jpmc}
  \end{icmlauthorlist}

  \icmlaffiliation{jpmc}{JPMorgan AI Research, NY, USA}
  \icmlaffiliation{comp}{Outsampler}

  \icmlcorrespondingauthor{Penghang Liu}{penghang.liu@jpmchase.com}

  \icmlkeywords{Machine Learning, ICML}

  \vskip 0.3in
]



\printAffiliationsAndNotice{\copyright 2026 JPMorgan Chase \& Co. All rights reserved. *Work done while at JPMorgan AI Research.}



\begin{abstract}
Large language models (LLMs) exhibit strong symbolic and compositional reasoning, yet they struggle with time series question answering as the data is typically transformed into an LLM-compatible modality, e.g., serialized text, plotted images, or compressed time series embeddings. Such conversions impose representation bottlenecks, often require cross-modal alignment or finetuning, and can exacerbate hallucination and knowledge leakage.
To address these limitations, we propose \textsc{TS-Agent}, an agentic, tool-grounded framework that uses LLMs strictly for iterative evidence-based reasoning, while delegating statistical and structural extraction to time series analytical tools operating on raw sequences.
Our framework solves time series tasks through an evidence-driven agentic process: (1) it alternates between thinking, tool execution, and observation in a ReAct-style loop, (2) records intermediate results in an explicit evidence log and corrects the reasoning trace via a self-refinement critic, and (3) enforces a final answer-verification step to prevent hallucinations and leakage.
Across four benchmarks spanning time series understanding and reasoning, \textsc{TS-Agent} matches or exceeds strong text-based, vision-based, and time-series language model baselines, with the largest gains on reasoning tasks where multimodal LLMs are prone to hallucination and knowledge leakage in zero-shot settings.
\end{abstract}

\begin{figure}[t]
    \centering
    \includegraphics[width=1.0\linewidth]{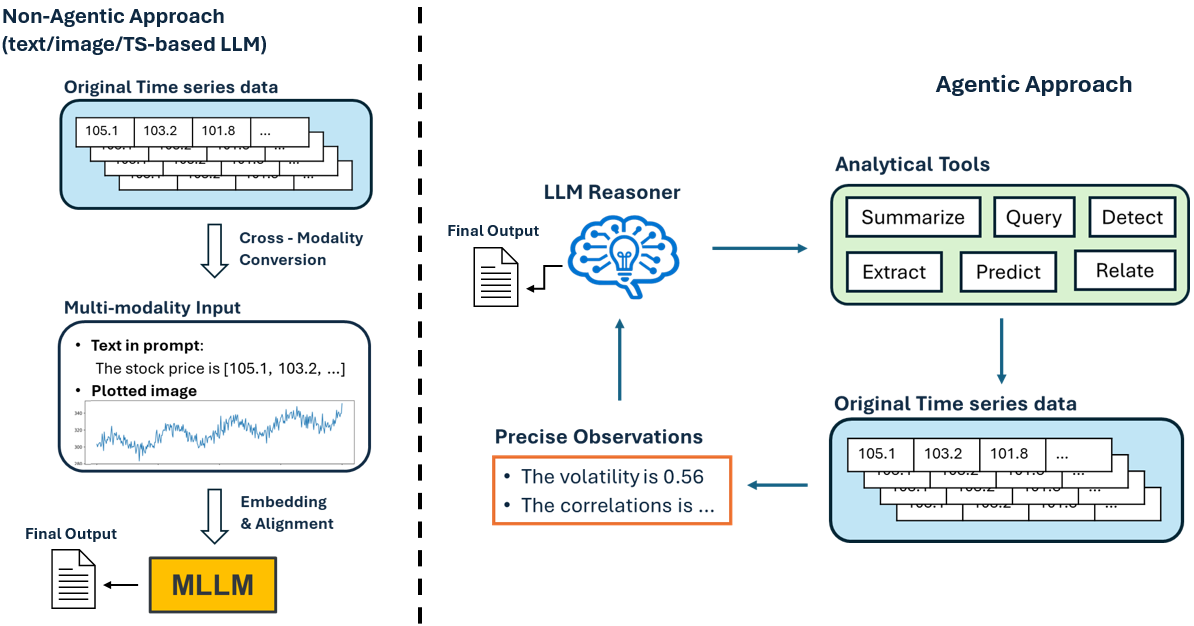}
    \caption{Comparison between agentic and non-agent approach for time series tasks.}
    \label{fig:Intro}
\end{figure}

\section{Introduction}
Time series data play a central role in many high-stakes domains such as finance, healthcare, climate science, and energy systems. Understanding and reasoning about these data is often crucial for decision-making.
For example, identifying whether a sudden change in a patient's heart rate signals a genuine medical risk, or determining whether a sharp movement in stock prices is driven by structural market factors or short-term noise. Such scenarios require more than simple forecasting, which demands factual understanding, mathematical reasoning, and the ability to uncover causal or correlational relationships in sequential data~\cite{shumway2017time,fulcher2017hctsa,ismail2019deep,wen2020time}.

Large language models (LLMs) and multi-modal LLMs (MLLMs) have recently demonstrated remarkable capabilities in reasoning and problem solving across a wide range of textual and mathematical tasks. Techniques such as chain-of-thought prompting and self-consistency have shown that LLMs can decompose complex questions into step-by-step reasoning trajectories and arrive at more reliable answers~\cite{wei2022chain,wang2022self,kojima2022large,zhou2023least}.
These developments suggest that LLMs could serve as general-purpose reasoning engines for time series tasks, and recent work has explored LLMs for forecasting, understanding, and time-series question answering by converting time series into an LLM-compatible modality, such as serialized text, plotted images for vision-language models, or compressed time-series embeddings~\cite{kong2025position, zhong2025time, gruver2023large, zhang2024large, kong2025time, xie2024chatts, fons2024evaluating, ye2024beyond}.

However, this prevailing ``convert-then-reason'' approach faces fundamental limitations. First, representing time series as text, images, or embeddings introduces a representation bottleneck: such conversion inevitably compresses the original signal which discards the fine-grained, precise information needed for accurate analysis. For instance, plotted images may convey global trends effectively yet obscure local statistical cues (e.g., brief outliers, short-lived regime changes, or window-specific variance) that are critical for many understanding questions. Second, reasoning over time series remains challenging for LLMs even when such representations are provided. Merrill et al.~\cite{merrill2024language} show that language models systematically underperform on zero-shot time series reasoning tasks such as etiological reasoning, context-aided forecasting, and question answering, attributing the gap to the mismatch between continuous quantitative inputs and token-based representations, difficulties with precise numerical manipulation, and the specialized inductive and causal reasoning required in time-series domains. Moreover, when asked to produce long-form explanations, LLMs may answer correctly without consulting the data due to knowledge leakage~\cite{merrill2024language}, or fabricate unsupported details as reasoning traces grow longer~\cite{xie2024chatts}. Taken together, these issues suggest that the central obstacle is not whether LLMs can \emph{reason} in general, but whether they can reliably \emph{access} and \emph{verify} the necessary evidence from the underlying time series.

Motivated by this gap, we argue that \emph{agentic} time-series intelligence is a better paradigm than forcing time series into an LLM/MLLM input modality (text/vision/embedding). We introduce \textsc{TS-Agent}, an AI agent that keeps the time series in its native numeric form and uses the LLM strictly where it is strongest: \emph{iterative}, \emph{evidence-based reasoning}, while delegating time-series understanding to domain-appropriate analytical functions that compute the required facts directly from the data. Concretely, \textsc{TS-Agent} addresses time-series QA through a ReAct-like process~\cite{yao2023react}: the LLM decomposes a question into iterative thinking process, where in each step it invokes an analytical actions to extract verifiable evidence from the raw sequence, and synthesizes a final answer grounded in computed results. This design yields three practical benefits. First, as shown in \cref{fig:Intro}, it eliminates the need for multi-modal alignment training to make an LLM ``perceive'' time series, because the LLM never has to parse long numeric sequences as tokens or interpret plots as pixels. Second, it avoids the information loss inherent to conversion-based pipelines by preserving direct access to the original signal, enabling precise local statistics and events (e.g., outlier) to be recovered when needed. Third, it improves reliability in complex reasoning settings via two explicit safeguards: a step-wise self-refinement critic that corrects the reasoning trace and a final answer verification mechanism that checks whether the conclusion is supported by the obtained evidence, thereby reducing hallucination and knowledge leakage.

We summarize our main contributions as follows:
\begin{itemize}
    \item We motivate an \emph{agentic} paradigm for time-series QA that reasons directly over native time series, instead of converting the signal into text, images, or embeddings for LLM/MLLM processing.
    \item We propose \textsc{TS-Agent}, a ReAct-style time-series agent that decomposes questions into iterative evidence-gathering steps over raw sequences. Our framework integrates a step-wise self-refinement module and a final answer verification mechanism to enhance time series reasoning capability and mitigate hallucination and knowledge leakage.
    \item We demonstrate strong empirical performance across four benchmarks spanning time series understanding and reasoning, outperforming text-based, vision-based and time series language model baselines, with particularly large gains on reasoning tasks and consistent improvements on local pattern understanding.
\end{itemize}

\section{Related Works}
\textbf{Time Series QA Benchmarks.}  
With the rapid advancement of LLMs, several works have proposed benchmarks and multimodal methods for handling time-series data questions~\cite{kong2025time,zhong2025time,gruver2023large,zhang2024large}. Among these, TimeSeriesExam~\cite{cai2024timeseriesexam} provides a large-scale multiple-choice exam across five categories (pattern recognition, noise understanding, anomaly detection, similarity, and causality), and Fons et al.~\cite{fons2024evaluating} introduce a taxonomy and benchmark for time-series feature understanding, retrieval, and arithmetic reasoning. Xie et al.~\cite{xie2024chatts} propose ChatTS, a multimodal LLM aligned with synthetic time series–text pairs, showing improvements on both forecasting and question answering. More recently, MMTS-Bench \cite{mmtsbench} provides a comprehensive dataset for multimodal time series understanding and reasoning. 
Chen et al.~\cite{chen2025mtbench} propose MTBench, a large-scale benchmark pairing time series with aligned textual narratives to evaluate cross-modal temporal reasoning, including forecasting, trend analysis, and news-driven question answering.
Despite these promising advances, Merrill et al.~\cite{merrill2024language} demonstrate that current LLMs still struggle on genuine reasoning tasks, often producing nearly random answers or achieving high accuracy even when the time-series input is removed, revealing that their performance frequently reflects memorized knowledge rather than actual data analysis.

\begin{figure*}[t]
    \centering
    \includegraphics[width=0.85\textwidth]{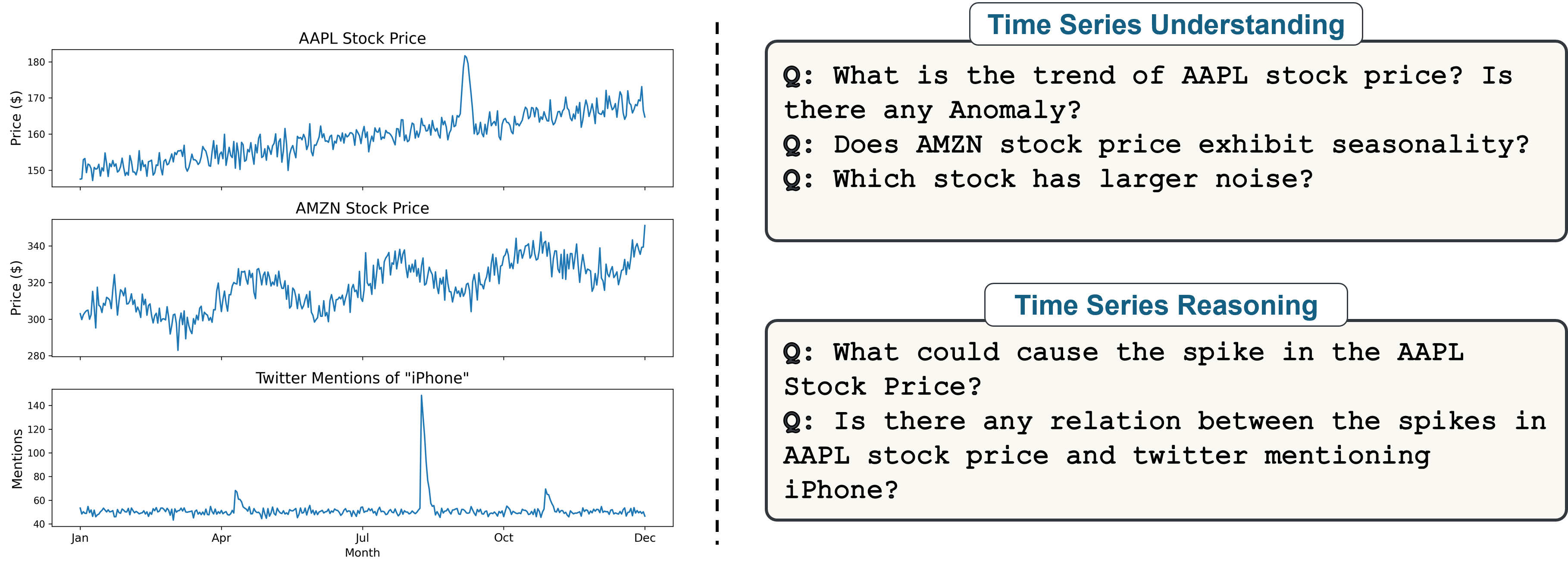}
    \caption{The two types of time series questions: understanding (top) and reasoning (bottom).}
    \label{fig:QA-Category}
\end{figure*}

\textbf{AI Agents for Reasoning.}  
General-purpose reasoning agents such as ReAct~\cite{yao2023react}, Reflexion~\cite{shinn2023reflexion}, and Self-Refine~\cite{madaan2023self} combine natural-language reasoning with tool use and self-correction. Recent surveys further highlight LLM-based agents as a promising paradigm for complex reasoning tasks across domains~\cite{wang2024survey,xi2025rise}. In the time series domain, Zhao et al.~\cite{zhao2025timeseriesscientist} introduce TimeSeriesScientist, an LLM-driven multi-agent framework that automates end-to-end time-series forecasting, rather than focusing on time-series question answering.
Ye et al.~\cite{ye2024beyond} propose TS-Reasoner, a domain-specific agent for electricity load forecasting. Their approach uses the LLM purely as a task decomposer that produces a complete plan executed by external programs. Our work differs in two key aspects: (i) we address general time-series reasoning across diverse domains and evaluate on non–domain-specific benchmarks; and (ii) our TS-Agent performs iterative ReAct-style reasoning, where each step is logged in an evidence log, reviewed by a critic, and verified by a final quality gate, ensuring transparency and verifiability rather than relying on a one-shot plan.

\section{Background: Time-Series Tasks}
Language models have been used to address time series questions from a wide spectrum, but the nature of these questions spans two fundamentally different categories. We distinguish between \emph{time-series understanding}, where the answer is a direct property or characteristic of the series, and \emph{time-series reasoning}, where the answer requires integrating multiple aspects of the series dynamics to infer higher-level conclusions. This distinction is critical: while many prior works evaluate LLMs on understanding tasks~\cite{cai2024timeseriesexam,fons2024evaluating}, recent studies show that reasoning remains a major challenge~\cite{merrill2024language}. Examples of the time series understanding and reasoning questions are given in \cref{fig:QA-Category}

\subsection{Time-Series Understanding}
Understanding tasks ask for explicit concepts or features of a time series. Formally, Given a multivariate time series $X \in \mathbb{R}^{d \times T}$ and a natural language question $q$, the target output $y$ is a \emph{characteristic} $c \in \mathcal{C}$ or \emph{statistic} $r \in \mathbb{R}$ that can be directly extracted from the data:
$$
y = f(X;\theta)
$$
using some task-specific function or model $f$ with parameter $\theta$. Typical examples include pattern recognition (trend, seasonality, anomaly, shifts), computation (variance, auto-correlation, forecasting), and comparison (similarity, correlation, causal relation).
Such problems can often be solved by a single analytic operator or by a specialized predictive/detection model, which correspond to the problems in TimeSeriesExam~\cite{cai2024timeseriesexam} and the time series feature understanding benchmark~\cite{fons2024evaluating}. While sometimes described as ``reasoning,'' these tasks in fact primarily require the ability to extract and identify well-defined properties from data.

\subsection{Time-Series Reasoning}
Reasoning tasks go beyond extracting properties to require logical inference about the dynamics and mechanics of the series. The answer itself is not a time-series concept but a conclusion derived from combining several such concepts. Formally, given $(X,q)$, the solution requires a composition
$$
y = \Phi\big(f_1(X), f_2(X), \dots, f_n(X)\big)
$$
where $\Phi$ integrates multiple intermediate properties into a higher-level inference. Some typical examples are: (1) Given temperature and electricity consumption series, predict the effect of a spike in temperature. Solving requires understanding causal relations (temperature $\rightarrow$ demand) and daily seasonality of consumption. (2) Given solar panel output over one month, identify a cloudy period. This demands integrating knowledge of diurnal cycles (peaks at noon, zero at night) with deviations in daily maxima.
(3) Choosing the best investment among several stocks, which requires analyzing trends, volatilities, correlations, and forecasts jointly.

Merrill et al.~\cite{merrill2024language} define several time series reasoning tasks and find that current LLMs perform near random or substantially below human baselines, underscoring that true reasoning about time series remains unsolved.


\section{Time Series Agent Framework}
\begin{figure*}[t]
    \centering
    \includegraphics[width=1.0\textwidth]{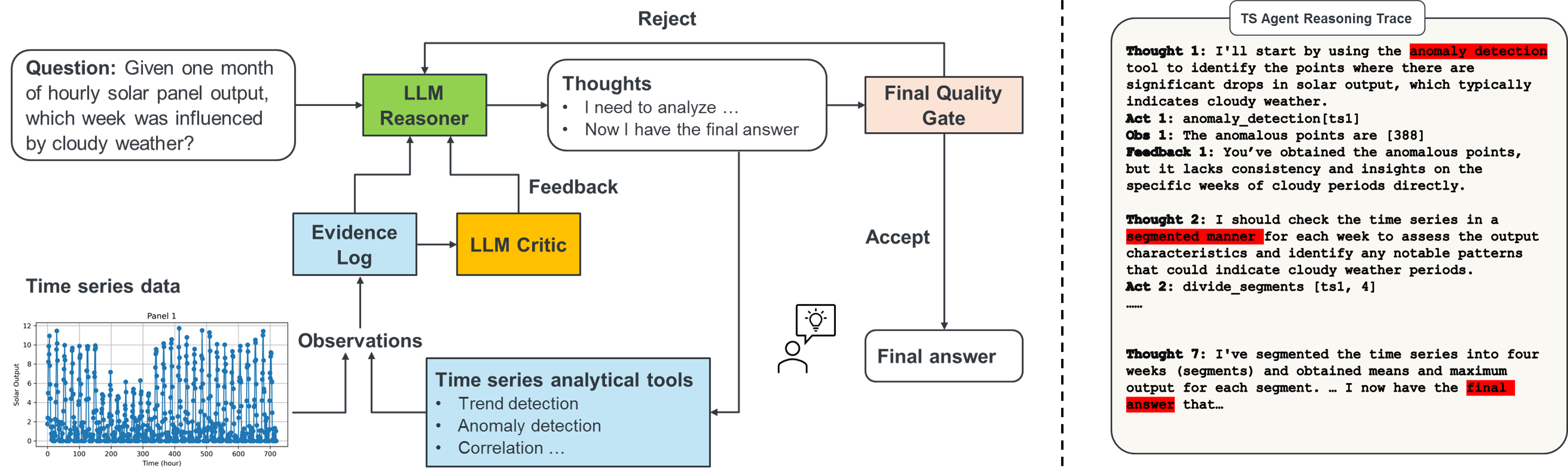}
    \caption{The TS Agent framework (left) and example of reasoning trace (right).}
    \label{fig:TS-Agent}
\end{figure*}

We design a time series agent that address both understanding and complex reasoning problems through an iterative process of reasoning, evidence gathering, and self-refinement. \Cref{fig:TS-Agent} presents the pipeline of TS Agent. Given a multivariate time series $X \in \mathbb{R}^{d \times T}$ and a natural language question $q$, the agent decomposes the problem into reasoning steps, calls analytical tools iteratively to extract evidence from $X$, and integrates observations into its chain of thought until reaching a final answer. In addition, we employ a critic to review each step for self-refinement, and a quality gate for final answer verification. This design ensures accurate quantitative analysis via tools, interpretability via step-by-step reasoning trace, and robustness through iterative self-refinement and final quality control, contrasting with non-agentic LLM approaches that directly map time series input into final answer without structured reasoning~\cite{cai2024timeseriesexam,zhong2025time}.

\subsection{Iterative LLM Reasoner}
At the core of our framework is an iterative LLM reasoner that drives the reasoning process through cycles of thought, action, and observation. Rather than producing a fixed plan upfront, the agent adaptively generates the next reasoning step only after incorporating the structured outcome of the previous step. This makes the agent responsive to evidence as it emerges, and robust against early mistakes. 

Formally, let $\mathcal{T}=\{t_1,\dots,t_M\}$ be the set of callable tools, and $\mathcal{L}_k=\{e_1,\dots,e_M\}$ be evidence log, i.e., the memory module that records of all accumulated observations produced during reasoning. The state of the agent at step $k$ consists of the query, a reasoning trace $\tau_{1:k-1}$ of past thoughts and actions, and the evidence log $\mathcal{L}_{k-1}$ containing accumulated observations. The LLM generates the next thought:
\[
\theta_k = f_{\mathrm{LLM}}(q, \tau_{1:k-1}, \mathcal{L}_{k-1}),
\]
where $\theta_k$ is a natural-language hypothesis or sub-goal.

Conditioned on this thought, the agent selects a tool $t_k \in \mathcal{T}$ and structured arguments $\alpha_k$, forming an \emph{action} $a_k = (t_k, \alpha_k)$.
Execution of this action produces a structured observation: $o_k = t_k(X; \alpha_k)$,
where $o_k$ may be numeric, categorical, or relational depending on the tool type.
The observation is saved in the evidence log, then the cycle proceeds with the updated context $(\tau_{1:k}, \mathcal{L}_k)$, ensuring that subsequent reasoning steps can build on all accumulated evidence. The trajectory continues until the LLM concludes that sufficient evidence has been gathered to answer the query or declares the problem undecidable. 

This iterative mechanism distinguishes our design from static planners: at each step, reasoning is grounded not only in the original question $q$ but also in the dynamically growing buffer of evidence $\mathcal{L}_k$. The explicit recording of intermediate observations improves interpretability and provides the foundation for later modules such as the critic and quality gate.

\subsection{Time Series Analytical Actions}
To support fine-grained and verifiable reasoning, \textsc{TS-Agent} is equipped with a library of callable analytical APIs that operate directly on native numeric time series and return structured outputs. Rather than forcing the signal into an LLM-compatible modality (text, plots, or embeddings), the agent queries the original sequence through a small set of time series analytical tools. Here we classify the tools into following types of \emph{human-like analysis actions}, where the complete tool list is in~\cref{app:tools}.

\paragraph{Summarize ($f_{\mathrm{sum}} : \mathcal{X} \times \Theta \to \mathbb{R}^m$).}
These tools provide concise numerical characterizations of a series or a specified window, including summary statistics (e.g., min/max/mean/std, quantiles and ranges), variability measures (e.g., rolling statistics and volatility), and task-relevant aggregates (e.g., return statistics). They supply quantitative evidence without making structural claims about the underlying dynamics.

\paragraph{Extract ($f_{\mathrm{ext}} : \mathcal{X} \times \Theta \to \mathcal{X}$).}
These tools extract part of the time series for subsequent local analysis, such as selecting channels, slicing a temporal interval, segmenting into subseries, or resampling to a desired granularity. By enabling targeted analysis at the appropriate resolution and context, \emph{Extract} tools prevent spurious conclusions driven by irrelevant portions of the signal.

\paragraph{Query ($f_{\mathrm{qry}} : \mathcal{X} \times \Theta \to \mathcal{Y}$).}
These tools retrieve precise values and locations from the series, returning scalar values, indices/timestamps, or collections of intervals (e.g., value at time $t$, argmax/argmin within a period, threshold-crossing times, or durations above/below a criterion). This category supports questions that require pinpointing specific time or value, which provides exact evidence for downstream reasoning.

\paragraph{Detect ($f_{\mathrm{det}} : \mathcal{X} \times \Theta \to \mathcal{C}$).}
These tools identify and classify interpretable phenomena in the time series, including trend, changepoints or regime shifts, anomalies/outliers, seasonality or periodicity, and dependence structure diagnostics (e.g., autocorrelation-based patterns). The outputs are symbolic descriptors, bridging raw numeric observations to higher-level evidence used by the reasoner.

\paragraph{Predict ($f_{\mathrm{pred}} : \mathcal{X} \times \Theta \to \hat{\mathcal{X}}$).}
\emph{Predict} operators produce forecasts over specified horizons (and sometimes uncertainty summaries), enabling questions about extrapolation and expected future behavior. Similar to \emph{Extract} tools, \emph{Forecast} outputs can be returned as intermediate data, which can be use for other analytical tools.

\paragraph{Relate ($f_{\mathrm{rel}} : \mathcal{X}^k \times \Theta \to \mathcal{Z}$).}
These tools quantify relationships among multiple time series, capturing association, similarity, lagged dependence, and directional predictability. Examples include correlation and distributional comparisons, similarity, cross-correlation, and causality tests (e.g., Granger-causality test).

\paragraph{On-demand operator synthesis.}
While the above actions cover the majority of benchmark questions, time series analysis is open-ended. When a query requires a specialized operation not present in the fixed library, \textsc{TS-Agent} synthesizes a build-on-demand tool from a natural-language specification by composing existing primitives, and then invoke it as part of the tool library.

Overall, this intent-centric tool interface ensures that each reasoning step is grounded in verifiable computations tied to the original numeric signal.

\subsection{Self-Refinement and Final Answer Verification}
Beyond the iterative reasoner, our framework introduces two complementary mechanisms that improve reliability and auditability: a step-wise \emph{critic} for \emph{self-refinement} inspired by~\cite{shinn2023reflexion,madaan2023self} and a \emph{quality gate} for final answer verification. In particular, we extract \emph{question intents}, which specifies the requriements in the question and what must be conducted on the data to answer the question. Formally,
\[
\phi_{\text{intent}}:\ q \;\mapsto\; 
\big(\underbrace{\texttt{task}}_{\text{type}},
\ \underbrace{\texttt{schema}}_{\text{output format}},
\ \underbrace{\mathcal{R}}_{\text{required predicates}}\big),
\]
where $\mathcal{R}=\{r_1,\dots,r_m\}$ is a set of predicates that must be supported by evidence, inferred from the keywords in the question. After $k$ steps, the evidence log $\mathcal{L}_k$ induces the covered set
\[
\begin{aligned}
\mathcal{C}_k
&= \{\, r\in\mathcal{R} \;:\; \exists\, e\in\mathcal{L}_k \text{ that verifies } r \,\},\\
\mathcal{G}_k
&= \mathcal{R}\setminus \mathcal{C}_k,
\end{aligned}
\]
as the current \emph{gap set}. Our critic and quality gate operate by comparing $\mathcal{G}_k$ against $\mathcal{R}$. In practice (\texttt{detect\_question\_intents()}), $\phi_{\text{intent}}$ is realized via transparent rule bundles: keyword/pattern matches map questions into a small set of intent classes (e.g., \texttt{MCQ\_anomaly\_location}, \texttt{trend\_direction}, \texttt{seasonality\_type}, \texttt{relation\_lagged}), each declaring its \texttt{schema} and $\mathcal{R}$. For example, a question of “In which part of the time series does the anomaly occur?” triggers \texttt{MCQ\_anomaly\_location} with
\[
\begin{aligned}
\mathcal{R}
=\{&\ \texttt{has\_anomaly},\\
   &\ \texttt{anomaly\_segment}\in\{\texttt{head},\texttt{middle},\texttt{end}\}\ \}.
\end{aligned}
\]
As tools run, the log populates $\mathcal{C}_k$ (e.g., anomaly detected in $[t_1,t_2]$); any unmapped predicate remains in $\mathcal{G}_k$ and drives subsequent steps.

\paragraph{Step-wise Critic.}  
After the execution of each action, we invoke the same LLM in a critic role to assess the new evidence in the context of the accumulated reasoning trace and the evidence log. Specifically, the critic reviews $(\theta_k, \mathcal{L}_k,\ \phi_{\text{intent}}(q))$ and checks:
(i) \emph{tool suitability}, whether the chosen operator aligns with the current sub-goal and question intent; (ii) \emph{output plausibility}, whether the observed values are consistent with prior evidence; and (iii) \emph{evidence sufficiency}, whether the log contains all predicates required to satisfy the question intent. 
Based on the review, the critic may raise suggestion for correction of a wrong usage, using a different tool, or gather missing evidence. This dynamic loop allows the agent to backtrack from weak evidence, avoid compounding errors, and iteratively converge toward stronger conclusions.

\paragraph{Final Answer Verification.}
When the agent proposes an answer $\hat{y}$, we pass it to a quality gate which enforces data-grounding before output. We formalize the gate as
\[
\Gamma(q,\mathcal{L}_k,\phi_{\text{intent}}(q),\hat{y}) \;\in\; \{\texttt{accept}\}\ \cup\ \{\texttt{reject}\}\times\Delta_k,
\]
where $\Delta_k$ is a set of concrete rejection reasons. The gate checks: (i) \emph{schema compliance}—$\hat{y}$ matches \texttt{schema} (e.g., belongs to the MCQ option set); and (ii) \emph{evidence sufficiency}—all required predicates are verified ($\mathcal{G}_k=\varnothing$) and no contradictions remain. If any check fails, the gate returns \texttt{reject} along with $\Delta_k$ (e.g., “missing \texttt{trend\_direction}”, “contradictory \texttt{seasonality} labels”), which is fed back to the reasoner to continue gathering evidence. The process repeats until acceptance or the step budget is exhausted; only then do we emit \texttt{AGENT\_FAILURE} with the unresolved $\Delta_k$. This is crucial because LLMs may sometimes answer correctly without consulting data due to knowledge leakage~\cite{merrill2024language} or fabricate unsupported details in longer chains~\cite{xie2024chatts}. By requiring acceptance from $\Gamma$, our framework ensures final outputs are justified solely by verifiable predicates extracted from the input time series and recorded in the evidence log.




\begin{table}[t]
\caption{Overview of benchmark datasets in our experiments. For TSandLang, we randomly sampled 4,000 samples from the original dataset containing 230k QAs.}
\centering
\small
\setlength{\tabcolsep}{5pt}
\begin{tabular}{l l c }
\toprule
\textbf{Dataset} & \textbf{Focus} & \textbf{\#QAs}  \\
\midrule
TSExam \cite{cai2024timeseriesexam}    & Understanding & 763     \\
Feature Und. \cite{fons2024evaluating}  & Understanding & 2{,}600  \\
TSandLang \cite{merrill2024language} & Reasoning     & 4{,}000 \\
MMTS-Bench \cite{mmtsbench} & Reasoning     & 2{,}424  \\
\bottomrule
\end{tabular}
\label{tab:datasets_overview}
\end{table}

\begin{table*}[t]
\centering
\caption{Comparison of model accuracy on TSExam dataset of time series understanding tasks~\cite{cai2024timeseriesexam}. Best results are in bold, second-best are underlined. * indicates results from prior work~\cite{cai2024timeseriesexam,ye2024beyond}.}
\begin{tabular}{lcccccc}
\toprule
\textbf{Model} & \textbf{Modality} & \textbf{Pattern Rec.} & \textbf{Noise Und.} & \textbf{Anomaly Det.} & \textbf{Similarity} & \textbf{Causality}  \\ \midrule
Phi-3.5* & Text & 0.44                & 0.24                & 0.25              & 0.41                & 0.25               \\
DeepSeek* & Text & 0.63                & 0.58                & 0.43              & \underline{0.66}                & 0.28               \\
GPT-4o-mini   & Text & 0.35                & 0.32               & 0.40              & 0.51                & 0.36               \\
GPT-4o   & Text & 0.38                & 0.39               & 0.29              & 0.46                & \underline{0.42}               \\
GPT-o1*   & Text & \underline{0.65}         & 0.62                & \textbf{0.58}              & 0.62                & 0.37               \\
Gemini-2.5   & Text & 0.29                & 0.32               & 0.26              & 0.35                & 0.30               \\
Gemini-2.5   & Vision & 0.51                & 0.54               & 0.46              & \textbf{0.69}                & 0.24               \\
Mistral-7B  & Text  & 0.29                 & 0.33               & 0.38              & 0.45                & 0.28               \\ 
ChatTS  & TS  & 0.37                & 0.39               &  0.36             &  0.53               &  0.30              \\ \midrule
\textbf{TS-Agent (Ours)}     & & \textbf{0.71}                & \underline{0.61}                & \underline{0.57}              & 0.57                & \textbf{0.55}     \\ \bottomrule         
\end{tabular}
\label{tab:TSExam}
\end{table*}

\section{Experiments and Results}
In this section, we evaluate our agent on time series understanding and reasoning benchmarks and compare against the state-of-the-art text-based, vision-based, and time series-based language models. \cref{tab:datasets_overview} presents the overview of the four datasets. We classify the benchmark datasets into understanding-focused and reasoning-focused based whether the majority of the questions requires time series reasoning. Across all experiments we use \texttt{gpt-4o-mini} to server as the backbone model for the reasoner and critic. Although this is not the strongest available model, our results demonstrate that even with a relatively lightweight backbone the proposed framework achieves competitive performance on understanding benchmarks and substantial gains on reasoning tasks.

\begin{table*}[t]
\centering
\caption{Performance (F1 scores) on time series feature understanding benchmark~\cite{fons2024evaluating}). Best results are in bold, second-best are underlined.}
\resizebox{\textwidth}{!}{
\begin{tabular}{lccccccccc}
\toprule
\textbf{Model} & \textbf{Modality} & \textbf{Trend} & \textbf{Seasonality} & \textbf{Anomalies} & \textbf{Volatility} & \textbf{Struct.~break} & \textbf{Fixed corr.} & \textbf{Lagged corr.} & \textbf{Changing corr.} \\
\midrule
GPT-4o-mini & Text & \textbf{0.98} & 0.17 & 0.31 & 0.13 & 0.43 & 0.77 & 0.34 & 0.49 \\
GPT-4o & Text & \textbf{0.98} & 0.17 & 0.47 & 0.10 & 0.35 & \textbf{0.82} & \underline{0.45} & \textbf{0.58} \\
Gemini-2.5 & Text & \textbf{0.98} & \underline{0.49} & \underline{0.57} & 0.19 & 0.48 & \textbf{0.82} & 0.09 & 0.43 \\
Gemini-2.5 & Vision & \textbf{0.98} & 0.23 & \underline{0.57} & \textbf{0.27} & 0.37 & \underline{0.80} & 0.28 & 0.51 \\
Mistral-7B & Text & 0.67 & 0.21 & 0.39 & 0.12 & 0.41 & 0.64 & 0.02 & 0.00 \\
Phi-3 & Text & 0.48 & 0.48 & 0.53 & 0.08 & 0.51 & 0.43 & 0.41 & 0.48 \\
ChatTS & TS & \textbf{0.98} & 0.33 & 0.45 & \underline{0.25} & \textbf{0.58} & 0.71 & 0.17 & \underline{0.53} \\
\midrule
\textbf{TS-Agent (Ours)} & & \textbf{0.98} & \textbf{0.58} & \textbf{0.87} & \underline{0.25} & \underline{0.55} & 0.53 & \textbf{0.55} & \underline{0.53}\\
\bottomrule
\end{tabular}
}
\label{tab:Taxonomy}
\end{table*}

\subsection{Time Series Understanding}
We first evaluate \textsc{TS-Agent} on time series understanding tasks using the TimeSeriesExam benchmark~\cite{cai2024timeseriesexam}, which contains 763 multiple-choice questions over five categories: pattern recognition, noise understanding, anomaly detection, similarity analysis, and causality analysis.
Table~\ref{tab:TSExam} compares \textsc{TS-Agent} with a range of strong baselines, including text-only LLMs, vision-language models, and a time-series language model (ChatTS). Overall, \textsc{TS-Agent} achieves the best average performance among the compared methods. The advantage is particularly clear on categories that require precise extraction from local or structured signal properties: \textsc{TS-Agent} attains the highest score on pattern recognition (0.71) and causality analysis (0.55), which has been reported as one of the hardest categories in TimeSeriesExam. Compared with LLM/MLLM baselines, we observe that vision inputs can be competitive on similarity-style questions where global shape cues are salient, while \textsc{TS-Agent} remains more reliable when questions depend on fine-grained numeric evidence (e.g., local events, controlled counterfactual patterns, or Granger-style relationships). These results support our central claim that keeping time series in their native form and extracting evidence via analytical operators can improve understanding without requiring any time-series-specific finetuning of the LLM.

We further evaluate understanding on the time series feature understanding benchmark of Fons et al.~\cite{fons2024evaluating}, which measures feature detection and classification for both univariate and multivariate series. Table~\ref{tab:Taxonomy} shows that \textsc{TS-Agent} achieves good performance across a broad set of characteristics. 
While our agent is not the best in recognizing volatility, it significantly outperforms the baseline on seasonality, anomalies, and structural break classification tasks which requires more detailed analysis of local time series patterns. For multivariate characteristics, \textsc{TS-Agent} is especially effective on lagged correlation, where identifying lead--lag structure requires explicit temporal dependence analysis. Overall, the results indicate that our agentic approach yields reliable understanding performance across both univariate and multivariate settings, while remaining training-free and directly grounded in the original time series.

Taken together, these understanding experiments show that \textsc{TS-Agent} can match or exceed strong text-based, vision-based, and time-series language model baselines, particularly on tasks that depend on local patterns and precise statistical evidence. We note that while certain modalities can excel on specific characteristic types, \textsc{TS-Agent} is the most consistent all-around approach which performs reliably well across the full spectrum of features..

\subsection{Time Series Reasoning}
Next, we evaluate \textsc{TS-Agent} on time-series reasoning tasks, where solving a question requires more than extracting a property from the input series. 
Table~\ref{tab:TSandLanguage} reports results on the TSandLanguage benchmark~\cite{merrill2024language} for one-series and two-series reasoning settings. A key challenge in this benchmark is that strong pretrained LLMs can achieve high scores even without access to the time series input, indicating substantial knowledge leakage and making it difficult to interpret raw accuracy as genuine reasoning~\cite{merrill2024language}. In contrast, \textsc{TS-Agent} is explicitly constrained to reason from computed evidence: it records all intermediate results in an evidence log, and the final answer verification mechanism rejects any unsupported conclusions. Even under this evidence-driven protocol, \textsc{TS-Agent} is still competitive on the one-series setting and achieves the strongest performance among all models on the two-series setting, which is particularly challenging because it requires integrating evidence across signals.

We further evaluate our model’s reasoning capability on MMTS-Bench, a comprehensive benchmark that covers five types of time-series reasoning: inductive, deductive, counterfactual, causal, and analogical reasoning. As shown in Table~\ref{tab:mmts-bench}, \textsc{TS-Agent} achieves the best overall performance among all compared texte-based, vision-based, and time series embedding-based baselines. \textsc{TS-Agent} remains strong across these subsets, suggesting that an agentic approach, i.e., iteratively gathering verifiable evidence from native time series and validating the final conclusion, is better suited for complex reasoning than pipelines that rely on lossy modality conversion and single-pass generation.




\begin{table}[t]
\centering
\caption{Comparison of model accuracy on TSandLang dataset of time series reasoning tasks~\cite{merrill2024language}. Best results are in bold, second-best are underlined. * indicates results from prior work. \citet{merrill2024language} showed that pretrained LLMs achieved 78--92\% accuracy even without time series data, indicating that the answers stem from knowledge leakage rather than time series reasoning. ** Human annotators achieved 67\% accuracy on the two time series tasks.}
\begin{tabular}{lccc}
\toprule
\textbf{Model} & \textbf{Modality} & \textbf{One TS}  & \textbf{Two TS}**  \\ \midrule
LLAMA-7B     & Text & 0.788* & 0.252* \\
LLAMA-13B    & Text & 0.825* & 0.258* \\
GPT-3.5      & Text & 0.882* & 0.274* \\
GPT-4        & Text & \textbf{0.923}* & 0.527* \\
GPT-4 & Vision & \underline{0.918}* & \underline{0.536}* \\
GPT-4o-mini & Text & 0.616& 0.441 \\
GPT-4o & Text & 0.798 & 0.431\\
Mistral-7B & Text & 0.663 & 0.410 \\
Gemini-2.5 & Text & 0.335 & 0.440\\
Gemini-2.5 & Vision & 0.271 & 0.483\\
ChatTS & TS & 0.713 & 0.493\\
\midrule
\textbf{TS-Agent (Ours)} &    & 0.805  & \textbf{0.573}   \\ \bottomrule
\end{tabular}
\label{tab:TSandLanguage}
\end{table}



\begin{table}[b]
\caption{Performance on MMTS-Bench dataset of time series reasoning tasks \cite{mmtsbench}. 
Best results are in bold, second-best are underlined. * indicates results from prior work.}
\centering
\small
\resizebox{\linewidth}{!}{
\begin{tabular}{l c c c c c c c}
\toprule
\textbf{Model} & \textbf{Modality} & \textbf{Average} &
\textbf{Base} & \textbf{InWild} & \textbf{Match} & \textbf{Align} \\
\midrule
Qwen2.5-7B*           & Text & 0.45 & \underline{0.33} & 0.44 & 0.40 & 0.69 \\
ChatTS*   & TS & 0.49 & \textbf{0.39} & 0.50 & 0.37 & 0.80 \\
ITFormer*             & TS & 0.31 & 0.31 & 0.33 & 0.24 & 0.29 \\
GPT-4o-mini          & Text & 0.43 & 0.17 & 0.52 & 0.44 & 0.85 \\
GPT-4o               & Text & \underline{0.52} & 0.22 & \underline{0.61} & \underline{0.51} & \underline{0.96} \\
Gemini-2.5     & Text & 0.45 & 0.18 & 0.53 & 0.39 & 0.93 \\
Gemini-2.5 & Vision & 0.51 & 0.27 & 0.58 & 0.44 & 0.92 \\
Mistral            & Text & 0.28 & 0.13   & 0.37   & 0.24   & 0.40      \\
\midrule
\textbf{TS-Agent(Ours)}     &        & \textbf{0.60} & 0.21 & \textbf{0.71} & \textbf{0.77} & \textbf{0.97} \\
\bottomrule
\end{tabular}
\label{tab:mmts-bench}
}
\end{table}

\subsection{Ablation Study}
\begin{figure}[t]
  \centering
  \begin{subfigure}[t]{0.49\linewidth}
    \centering
    \includegraphics[width=\linewidth]{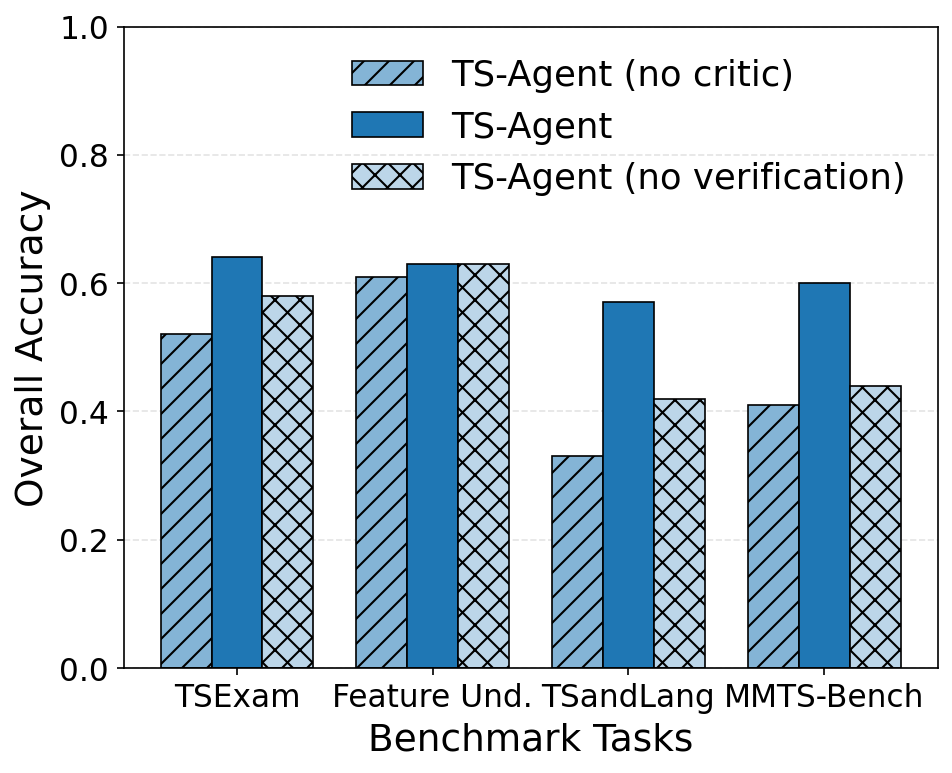}
    \caption{Overall accuracy.}
  \end{subfigure}\hfill
  \begin{subfigure}[t]{0.49\linewidth}
    \centering
    \includegraphics[width=\linewidth]{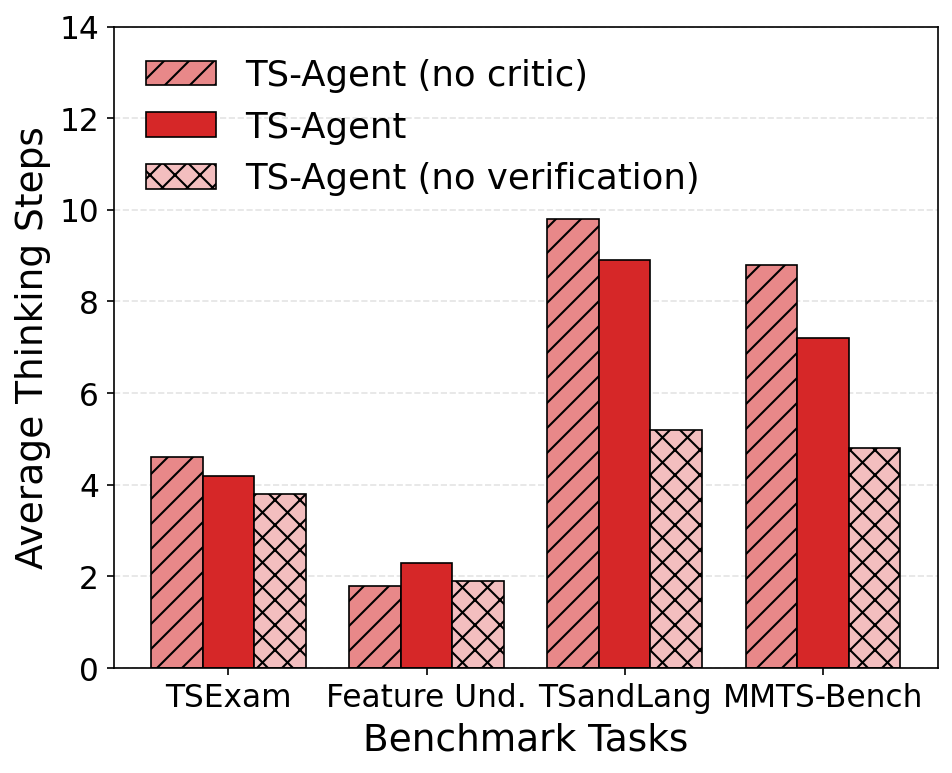}
    \caption{Average thinking steps.}
  \end{subfigure}
  \caption{The overall accuracy and average thinking steps of TS-Agent and the ablation models on all benchmark datasets.}
  \label{fig:ablation}
\end{figure}

As shown in the previous sections, \textsc{TS-Agent} relies on two key components: a step-wise LLM critic for self-refinement and a final answer verification mechanism. To study their importance, we evaluate two ablation variants: \textsc{TS-Agent} (no critic), where the critic is removed, and \textsc{TS-Agent} (no verification), where the final answer verification is disabled. We report overall accuracy on four benchmark datasets, together with the average number of think--act--observe iterations taken before the agent outputs an answer.
\cref{fig:ablation} summarizes the results on two understanding benchmarks (TimeSeriesExam, Feature Und.) and two reasoning benchmarks (TSandLang, MMTS-Bench). As expected, the agent uses fewer thinking steps on the two understanding benchmarks than on the two reasoning benchmarks, since understanding questions typically require fewer steps of evidence gathering and composition to reach a correct answer. 

Removing the critic leads to a slight accuracy drop on understanding tasks, which becomes more significant on the complex reasoning benchmarks. Meanwhile, the average number of thinking iterations increases, especially on TSandLang and MMTS-Bench. This suggests that critic feedback helps the reasoner stay on track and avoid circling around to get to the right tracks, enabling it to reach the correct line of reasoning with fewer iterations.

Removing the final answer verification also reduces accuracy, with a larger impact on the reasoning benchmarks. In contrast to the no-critic setting, disabling verification reduces the average number of thinking iterations substantially, especially on TSandLang and MMTS-Bench. This indicates that without final answer verification, the agent often stops too early before gathering sufficient evidence, and is more likely to output an unsupported conclusion (or a guess) in challenging multi-step reasoning problems. The verification mechanism therefore acts as a quality gate that forces the agent to continue reasoning until its conclusion is adequately supported.

In summary, the ablation results confirm that both the critic and the final answer verification are essential for \textsc{TS-Agent}'s performance, and their contributions are more substantial on complex time series reasoning tasks, where solving the problem is inherently harder and often requires longer reasoning traces.

\section{Conclusion}
In this paper, we present \textsc{TS-Agent}, an agent framework for time series understanding and reasoning that avoids converting time series into another input modalities (text, plots, or embeddings).
\textsc{TS-Agent} keeps the signal in its native numeric form, uses the LLM strictly for iterative evidence-based reasoning, and relies on human-like analytical actions (Summarize, Extract, Query, Detect, Predict, and Relate) to obtain verifiable facts from the data. To improve reliability, the framework includes a step-wise self-refinement critic and a final answer verification mechanism to mitigate hallucination and knowledge leakage.

Across four benchmarks spanning time series understanding and reasoning, \textsc{TS-Agent} outperforms text-based, vision-based, and time-series language model baselines, especially on reasoning tasks. These results highlight the effectiveness of agentic, evidence-grounded LLM systems for time series analysis without time-series-specific finetuning or multimodal alignment. 

\section*{Disclaimer}
This paper was prepared for informational purposes by the CDAO group of JPMorgan Chase \& Co and its affiliates (``J.P. Morgan'') and is not a product of the Research Department of J.P. Morgan. J.P. Morgan makes no representation and warranty whatsoever and disclaims all liability, for the completeness, accuracy or reliability of the information contained herein. This document is not intended as investment research or investment advice, or a recommendation, offer or solicitation for the purchase or sale of any security, financial instrument, financial product or service, or to be used in any way for evaluating the merits of participating in any transaction, and shall not constitute a solicitation under any jurisdiction or to any person, if such solicitation under such jurisdiction or to such person would be unlawful.

\bibliography{main}
\bibliographystyle{icml2026}

\newpage
\appendix
\onecolumn

\section{Benchmark Datasets}
Here we provide detailed information on the benchmark datasets used in our evaluation. We cover four recent benchmarks that target complementary aspects of time series question answering: \emph{TimeSeriesExam}~\cite{cai2024timeseriesexam}, the \emph{time series feature understanding benchmark}~\cite{fons2024evaluating}, the \emph{time series reasoning benchmark}~\cite{merrill2024language}, and \emph{MMTS-Bench}~\cite{mmtsbench}. Together they span tasks from low-level feature understanding to multi-step reasoning, including settings that stress multimodal alignment and in-the-wild generalization.

\textbf{TimeSeriesExam.}
TimeSeriesExam is designed as a comprehensive multiple-choice exam for testing time series understanding. It consists of 763 questions generated from 104 templates and refined with item response theory (IRT) to balance difficulty. The benchmark is divided into five major categories: pattern recognition, noise understanding, anomaly detection, similarity analysis, and causality analysis. Each category is further divided into subcategories such as trend recognition, stationarity detection, or Granger causality. All questions are grounded in synthetic or counterfactual series with known ground truth.

\textbf{Time Series Feature Understanding Benchmark.}
The feature understanding benchmark focuses on more fine-grained feature identification and classification tasks. Each task type includes 200 multiple-choice questions constructed from synthetic time series with controlled attributes. The benchmark is divided into \emph{univariate} and \emph{multivariate} settings. Univariate tasks include detection and classification of trend, seasonality, anomalies, volatility, and structural breaks. Multivariate tasks include identifying fixed, lagged, or changing correlation structures. 

\textbf{Time Series Reasoning Benchmark.}
The reasoning benchmark proposed by Merrill et al.~\cite{merrill2024language} evaluates whether LLMs can reason about time series in a zero-shot manner. It covers three families of tasks: \emph{etiological reasoning} (choosing the most plausible generative scenario or description for a given series), \emph{time-series question answering} (answering factual questions based on one or two input series), and \emph{context-aided forecasting} (using auxiliary textual context in addition to the series to forecast future behavior). The dataset is large-scale, consisting of thousands of question–series pairs, and has been used to demonstrate that LLMs often succeed without even accessing the series data, raising concerns about knowledge leakage and memorization.

\textbf{MMTS-Bench.}
MMTS-Bench is a multi-modal time series evaluation benchmark built around a hierarchical taxonomy that ranges from basic perception to advanced reasoning. It contains 2,424 time-series question--answer pairs organized into four subsets. \emph{Base} uses synthetic time series to assess structural awareness and feature analysis. The remaining subsets are constructed from real-world time series across multiple domains: \emph{InWild} emphasizes feature analysis and temporal reasoning under realistic conditions; \emph{Match} evaluates sequence similarity and morphological correspondence; and \emph{Align} tests bidirectional conversion between time series and natural language, probing cross-modal semantic understanding. This design provides a broad and challenging testbed that complements the more targeted understanding and reasoning benchmarks above.

\begin{table}[t]
\centering
\small
\renewcommand{\arraystretch}{1.3} 
\setlength{\tabcolsep}{6pt}
\begin{tabular}{l|l l}
\toprule
\textbf{Task} & \textbf{Category} & \textbf{Example Question} \\
\midrule
\multicolumn{3}{l}{\textbf{TimeSeriesExam}~\cite{cai2024timeseriesexam}} \\
\hline
Pattern recognition & Trend recognition & Does the series show an upward trend? \\
 & Stationarity detection & Is the series stationary over time? \\
 & Regime switching & Does the series switch between regimes? \\
 & Cycle recognition & Does the series exhibit repeating cycles? \\
 & AR/MA recognition & Which ARMA model fits the series? \\
\hline
Noise understanding & White noise recognition & Is the series indistinguishable from white noise? \\
 & Red noise recognition & Does the series resemble red noise? \\
 & Signal-to-noise ratio understanding & Is the signal-to-noise ratio high or low? \\
\hline
Anomaly detection & General anomaly detection & Where does an anomaly occur in the series? \\
\hline
Similarity analysis & Shape similarity & Which series is most similar in shape? \\
 & Distributional similarity & Which series has a similar distribution? \\
\hline
Causality analysis & Granger causality & Does series A Granger-cause series B? \\
\midrule
\multicolumn{3}{l}{\textbf{Feature Understanding Benchmark}~\cite{fons2024evaluating}} \\
\hline
Univariate detection & Trend & Does the series increase or decrease? \\
 & Seasonality & Does the series show seasonality? \\
 & Anomalies & Does the series contain anomalies? \\
 & Volatility & Is the volatility high or low? \\
 & Structural break & Is there a structural break? \\
\hline
Univariate classification & Trend & Which type of trend is present? \\
 & Seasonality & Which seasonal pattern is present? \\
 & Anomalies & Which anomaly type is present? \\
 & Volatility & What is the volatility level? \\
 & Structural break & What type of break occurred? \\
\hline
Multivariate & Fixed correlation & Are the two series correlated? \\
 & Lagged correlation & Does one series lag the other? \\
 & Changing correlation & Does correlation change over time? \\
\midrule
\multicolumn{3}{l}{\textbf{TSandLanguage}~\cite{merrill2024language}} \\
\hline
Etiological reasoning & Generative scenario & Which description best explains the series? \\
TS QA (one series) & Question answering & At what time does the series peak? \\
TS QA (two series) & Comparative reasoning & Which series reacts first to a shock? \\
Context-aided forecasting & Forecasting & Given context, what is the next value? \\
\midrule
\multicolumn{3}{l}{\textbf{MMTS-Bench}~\cite{mmtsbench}} \\
\hline
Feature Analysis & Trend Analysis & Identifies long-term directional patterns and trend strength. \\
 & Seasonality Analysis & Captures seasonal patterns and seasonality strength. \\
 & Noise Analysis & Distinguishes random fluctuations from signal components. \\
 & Volatility Analysis & Quantifies temporal variability and instability. \\
 & Basic Analysis & Computes fundamental statistics. \\
\hline
Temporal Reasoning & Deductive Reasoning & Applies general rules to infer properties of specific intervals. \\
 & Inductive Reasoning & Generalizes characteristics from observed sequences. \\
 & Causal Reasoning & Identifies causal or lead-lag relationships between series. \\
 & Analogical Reasoning & Analogical Reasoning
Infers similarity by comparing temporal patterns. \\
 & Counterfactual Reasoning & Predicts outcomes under hypothetical changes. \\
\bottomrule
\end{tabular}
\vspace{1ex}
\caption{Overview of tasks in the bechmark dataset\cite{cai2024timeseriesexam, fons2024evaluating, merrill2024language}.}
\label{tab:benchmarks}
\end{table}

\section{Detailed Description of Time Series Tools} \label{app:tools}
Below we provide concise, implementation-oriented notes on the analytical tools used by \textsc{TS-Agent}. Tools are grouped by analytical \emph{actions} as in the main text: \textbf{Summarize}, \textbf{Extract}, \textbf{Query}, \textbf{Detect}, \textbf{Predict}, and \textbf{Relate}. Each tool takes typed arguments, operates on raw numeric series, and returns structured outputs (e.g., numeric value(s), categorical label(s), index ranges), which are written to the evidence log with provenance.

\medskip
\noindent\textbf{Summarize.}

\noindent-- \texttt{series\_info} (\texttt{name}). Return basic metadata: length $T$, dimension $d$, channel names/indices, sampling interval (if available), missingness stats.

\noindent-- \texttt{summary\_stats} (\texttt{name}, \texttt{range}=\texttt{None}). Mean, std, min, max (optionally over a window/segment).

\noindent-- \texttt{datarange\_value} (\texttt{name}, \texttt{start}, \texttt{end}, \texttt{stat}=\texttt{mean}|\,\texttt{sum}|\,\texttt{max}|\,\texttt{min}). Compute a statistic over a specified window.

\noindent-- \texttt{rolling\_stat} (\texttt{name}, \texttt{stat}, \texttt{window}, \texttt{step}=1). Rolling mean/std/quantile with window size and step.

\noindent-- \texttt{quantile\_value} (\texttt{name}, \texttt{q}). Empirical quantile at level $q\in(0,1)$.

\noindent-- \texttt{return\_calc} (\texttt{name}, \texttt{t1}, \texttt{t2}, \texttt{kind}=\texttt{pct}|\,\texttt{diff}). Simple or percentage return between two times.

\noindent-- \texttt{autocorr} (\texttt{name}, \texttt{lag}). Autocorrelation at a specified lag.

\noindent-- \texttt{volatility} (\texttt{name}, \texttt{window}). Windowed volatility (e.g., std of differences or returns).

\medskip
\noindent\textbf{Extract.}

\noindent-- \texttt{slice\_series} (\texttt{name}, \texttt{start}, \texttt{end}). Extract a subsequence by timestamps/indices and return the sliced series.

\noindent-- \texttt{segment\_series} (\texttt{name}, \texttt{k} \textit{or} \texttt{lengths}). Partition a series into $k$ equal segments or a user-specified list of segment lengths; return segments with boundaries.

\noindent-- \texttt{resample\_series} (\texttt{name}, \texttt{interval} / \texttt{rate}, \texttt{method}=\texttt{mean}/\,\texttt{sum}/\,\texttt{last}). Temporal down/up-sampling with aggregation.

\noindent-- \texttt{select\_channel} (\texttt{name}, \texttt{names} \textit{or} \texttt{indices}). Extract one or more channels from a multivariate series.

\noindent-- \texttt{normalize\_series} (\texttt{name}, \texttt{method}=\texttt{z-score}/\,\texttt{minmax}, \texttt{ref\_window}=\texttt{None}). Normalize values globally or over a reference window.

\medskip
\noindent\textbf{Query.}

\noindent-- \texttt{datapoint\_value} (\texttt{name}, \texttt{index} \textit{or} \texttt{timestamp}). Return the value at a specific time.

\noindent-- \texttt{argmin\_argmax} (\texttt{name}, \texttt{range}=\texttt{None}, \texttt{which}=\texttt{min}|\,\texttt{max}). Return extreme value and its index/timestamp within a range.

\noindent-- \texttt{threshold\_crossings} (\texttt{name}, \texttt{threshold}, \texttt{direction}=\texttt{above}|\,\texttt{below}, \texttt{range}=\texttt{None}). Return indices/timestamps where the series crosses a threshold.

\noindent-- \texttt{duration\_above\_below} (\texttt{name}, \texttt{threshold}, \texttt{direction}=\texttt{above}|\,\texttt{below}, \texttt{range}=\texttt{None}). Return total duration (or spans) satisfying a criterion.

\medskip
\noindent\textbf{Detect.}

\noindent-- \texttt{trend\_classifier} (\texttt{name}, \texttt{window}=\texttt{None}). Classify global/segment trend as \{\texttt{up}, \texttt{down}, \texttt{flat}\}.

\noindent-- \texttt{seasonality\_detector} (\texttt{name}, \texttt{max\_period}). Detect periodicity; return period estimate and a strength label.

\noindent-- \texttt{change\_point\_detector} (\texttt{name}, \texttt{penalty}/\texttt{n\_cp}). Identify structural breaks in mean/variance; return change-point indices.

\noindent-- \texttt{anomaly\_classifier} (\texttt{name}, \texttt{threshold}, \texttt{window}=\texttt{None}). Flag anomalies; return indices/spans and a brief type label (\texttt{spike}, \texttt{dip}, \texttt{level\,shift}).

\noindent-- \texttt{spike\_detector} (\texttt{name}, \texttt{threshold}, \texttt{min\_sep}). Locate isolated spikes/dips with minimum separation.

\noindent-- \texttt{noise\_profile} (\texttt{name}, \texttt{window}=\texttt{None}). Qualitative noise label (e.g., \texttt{white}/\texttt{red}) based on simple tests.

\noindent-- \texttt{stationarity\_test} (\texttt{name}, \texttt{test}=\texttt{adf}|\,\texttt{kpss}). Return \{\texttt{stationary}, \texttt{nonstationary}\} plus test statistic.

\medskip
\noindent\textbf{Predict.}

\noindent-- \texttt{forecast\_series} (\texttt{name}, \texttt{horizon}, \texttt{method}=\texttt{naive}|\,\texttt{seasonal}|\,\texttt{ar}/\texttt{other}). Produce a forecast over a specified horizon; return predicted values (and optional uncertainty summaries when available).

\medskip
\noindent\textbf{Relate.}

\noindent-- \texttt{corr\_relation} (\texttt{name1}, \texttt{name2}, \texttt{lag}=0, \texttt{method}=\texttt{pearson}|\,\texttt{spearman}). Correlation (optionally lagged).

\noindent-- \texttt{cross\_correlation} (\texttt{name1}, \texttt{name2}, \texttt{max\_lag}). Cross-correlation function and best lag.

\noindent-- \texttt{dtw\_distance} (\texttt{name1}, \texttt{name2}). Dynamic time warping distance; lower implies higher similarity.

\noindent-- \texttt{shape\_similarity} (\texttt{name1}, \texttt{name2}, \texttt{norm}=\texttt{zscore}). Scale-invariant shape comparison score.

\noindent-- \texttt{granger\_causality} (\texttt{name1}, \texttt{name2}, \texttt{maxlag}). Test whether $X_1$ Granger-causes $X_2$; return $p$-value and decision.

\medskip
\noindent\textbf{On-demand tool synthesis.}

\noindent-- \texttt{custom\_tool} (\texttt{prompt}). Generate a task-specific tool from a natural-language description (e.g., ``volatility-adjusted moving average''); returns a callable signature and a short schema.

\medskip
\noindent\textbf{Notes.}
(1) All tools accept \texttt{name}/\texttt{name1}/\texttt{name2} to identify the target series/channel(s).
(2) Many tools support an optional \texttt{window}/\texttt{range} to localize computation.
(3) Outputs are typed and include minimal diagnostics (e.g., indices, confidence/test statistics) to support self-refinement and final answer verification.

\section{Prompts for LLM Modules}
In this section we provide the prompts used for the two language model modules in our framework. The LLM Reasoner prompt guides the model to follow a structured reason-act-observe procedure, invoking time series tools when necessary and producing a final answer in the required format. The LLM Critic prompt specifies how the critic model evaluates each reasoning step by checking tool appropriateness, the plausibility of outputs, and the sufficiency of accumulated evidence. Both prompts are shown below.

\begin{tcolorbox}[colback=gray!5,colframe=black!50,title=Reasoner Prompt]
You are tasked with answering the following question. You have access to a set of tools:

\{tool\_descs\}

These tools allow you to access and analyze the time series. When producing the Final Answer, you must follow the required format and respond in plain English text (no Markdown). The final output must be written exactly as: ``Final Answer: the final answer to the original input question''.

Follow this reasoning format:

Question: the input question to solve  

Thought: you should always think about what to do

Action: the tool to use, chosen from [\{tool\_names\}]  

Action Input: the arguments to the tool  

Observation: the tool’s returned result  

Feedback: the critic’s feedback  

......(The Thought/Action/Action Input/Observation/Feedback block may be repeated until you get final answer.)

Thought: I now know the final answer  

Final Answer: the final answer to the original input question

Begin!  

Question: \{query\}
\end{tcolorbox}

\begin{tcolorbox}[colback=gray!5,colframe=black!50,title=Critic Prompt]
You are the critic reviewing the reasoning process. You are given the current observation and the accumulated evidence log. Your task is to assess:

(i) Tool suitability: does the chosen operator match the sub-goal and the question intent?  

(ii) Output plausibility: are the observed values consistent with prior evidence and expectations?  

(iii) Evidence sufficiency: does the log contain enough information to fully satisfy the question intent?

Provide concise feedback indicating whether the reasoning is appropriate, highlight potential issues, and suggest corrections if necessary.
\end{tcolorbox}

\section{Case Study: Reasoning Trace on Cloudy Period Detection}
To illustrate how the agent performs iterative reasoning with self-refinement, we present a representative example of its reasoning trace on a time series analysis task. The question asks the agent to identify the weeks during which solar panel output is reduced due to cloudy weather. The input consists of two month-long hourly solar output series with daily seasonality (peaks at noon, zero output at night), along with periods of cloudy weather.

The agent’s initial intuition is to treat cloudy conditions as anomalies, since cloudy weather reduces output abruptly. Therefore, the first attempt is to call the \texttt{anomaly\_classifier} tool. However, as shown in the first reasoning trace box, the agent misuses the anomaly detection tool by providing two time series simultaneously, which causes an error. The critic provides feedback about the expected input format, leading the agent to correct its tool usage.

After correcting the usage, the agent applies anomaly detection separately to each series. As illustrated in the second reasoning trace box, this produces very limited and misleading results: the anomaly detector does not capture cloudy periods well because cloudy spans are continuous intervals rather than isolated anomalous points. The critic suggests that segmentation would be a more appropriate approach.

In the third reasoning trace box, the agent revises its plan: it segments each monthly time series into four weeks and compares the average outputs across weeks. This approach directly aligns with the question’s intent, as cloudy weeks can be identified by sustained reductions in mean or maximum solar output. By following this refined reasoning path, the agent arrives at the correct final answer.

This case study highlights how the agent uses step-by-step reasoning and critic feedback to refine its approach: starting with an intuitive but unsuitable choice of anomaly detection, correcting tool misuse, and finally adopting segmentation as the appropriate method. Such iterative refinement enables the agent to better align its tool selection with the characteristics of the data and the requirements of the question.

\begin{tcolorbox}[colback=gray!5,colframe=black!50,title=Example Question]
\textbf{Question.}  
You are a time series analysis expert. The time series represents hourly solar output from a panel over a month, influenced by a period of cloudy weather. Please analyze the time series features and answer the following questions:  

How does the duration of cloudy periods within the month compare between the first and second time series?  

Please strictly follow the output format as:  

\texttt{|the first time series has cloudy periods for the (select all that apply from first, second, third, fourth) week, whereas the second time series are cloudy for the (select all that apply from first, second, third, fourth) week.|}  

Then briefly explain your answer.

\begin{center}
\includegraphics[width=0.45\linewidth]{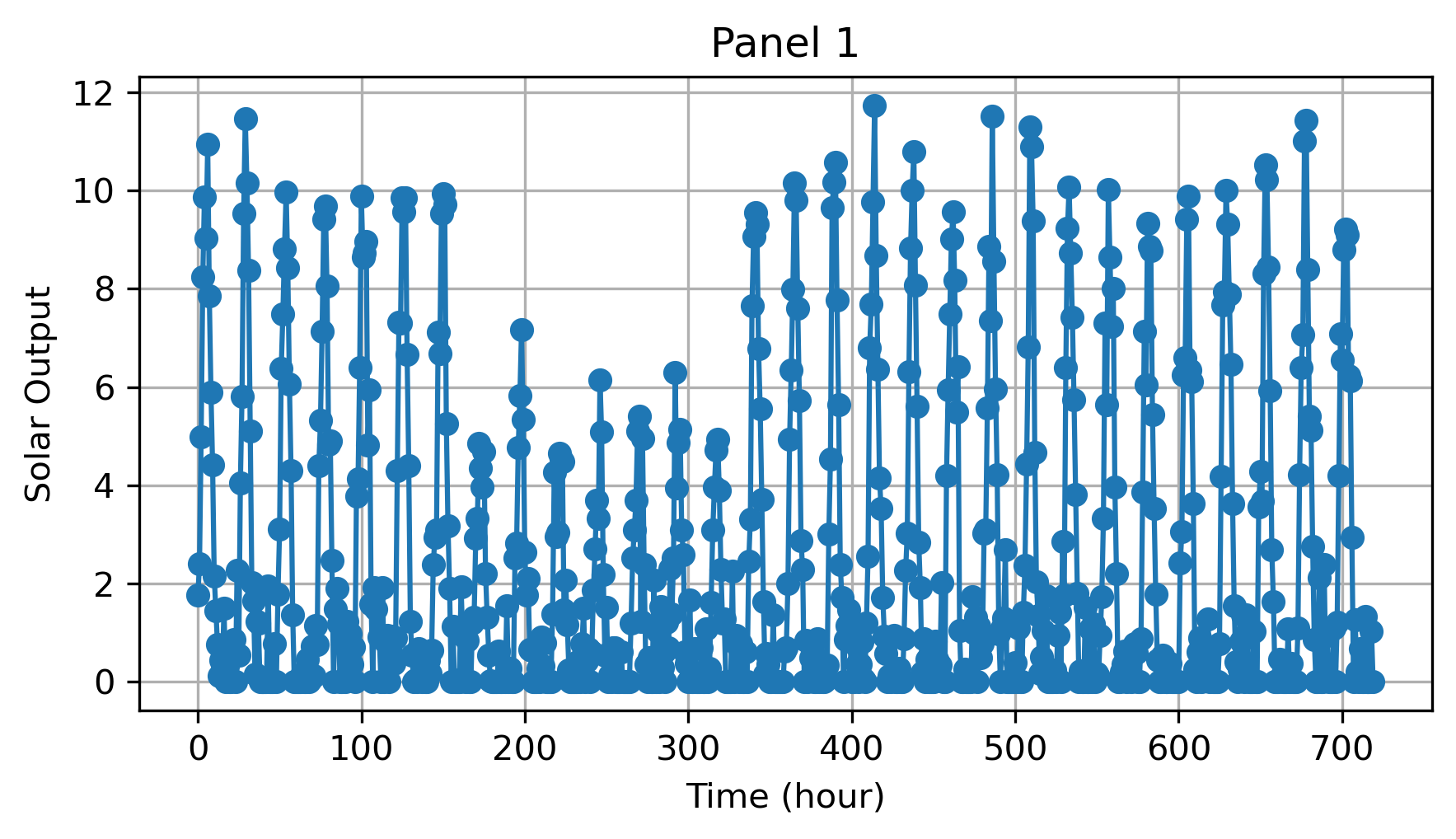}\hspace{0.5cm}
\includegraphics[width=0.45\linewidth]{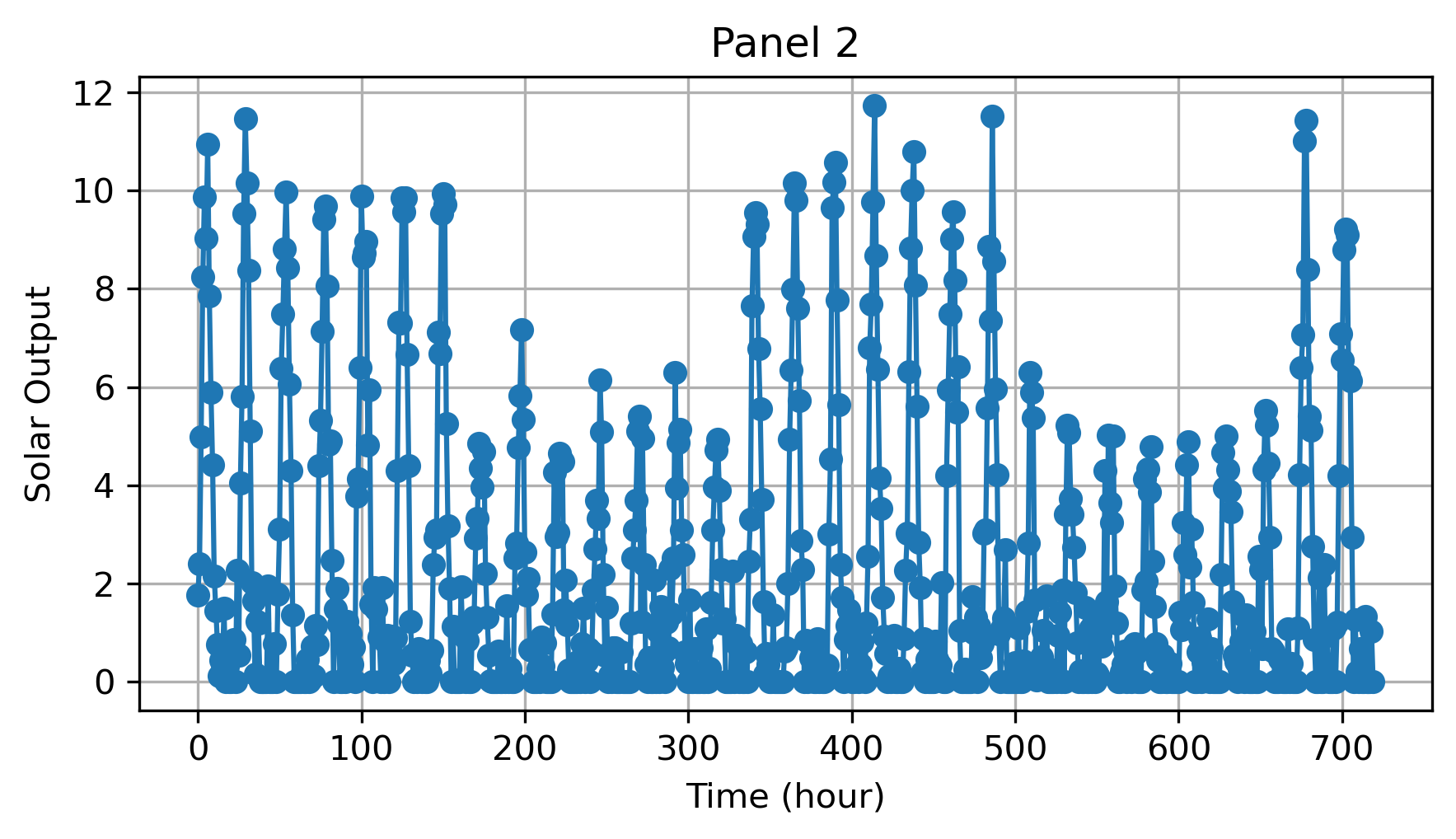}
\end{center}
\end{tcolorbox}

\begin{tcolorbox}[colback=gray!5,colframe=black!50,title=Self-refinement: Error Correction]
\texttt{\textbf{Thought:} To analyze the duration of cloudy periods in the two time series, I first need to determine the characteristics of the time series, particularly focusing on anomalies related to cloudy weather. I'll start by using the \colorbox{yellow!40}{anomaly detection tool} to identify the points where there are significant drops in solar output, which typically indicates cloudy weather.}

\vspace{0.3em}
\texttt{\textbf{Action:} anomaly\_classifier, Action Input: \{\colorbox{red!40}{Solar panel 1,} \colorbox{red!40}{Solar panel 2}\}}

\vspace{0.3em}
\texttt{\textbf{Observation:} Error when calling anomaly\_classifier: x must be 1-D}

\vspace{0.3em}
\texttt{\textbf{Feedback:} Error calling anomaly\_classifier, Input: \{Solar panel 1, Solar panel 2\}, It seems there was an error when trying to perform anomaly detection \colorbox{red!40}{across multiple time series.} The correct usage i \{'parameters':[ \{'name': 'name', 'required': True\}, \{'name':'threshold', 'required': False\}].}
\end{tcolorbox}

\begin{tcolorbox}[colback=gray!5,colframe=black!50,title=Reasoning Trace: Corrected Tool Usage]
\texttt{\textbf{Thought:} I'll try performing the \colorbox{yellow!40}{anomaly detection on the single} \colorbox{yellow!40}{time series} "Solar panel 1" to analyze the cloudy periods more effectively.}

\texttt{\textbf{Action:} anomaly\_classifier, Action Input: \{Solar panel 1\}}

\texttt{...}

\texttt{\textbf{Observation:} For Solar panel 1, no anomaly is detected. For Solar panel 2, the anomalous points are [388].}

\texttt{\textbf{Feedback:} The anomaly detection indicated that there are no cloudy periods detected in Solar panel 1, while Solar panel 2 has an anomalous point. To gain a clearer understanding of how cloudy periods compare between the two time series, it would be prudent to \colorbox{red!40}{segment the time series into weeks and analyze the output} to determine the presence of reduced output due to cloudy weather in each week.}

\texttt{\textbf{Thought:} I need to divide the time series for "Solar panel 1" into four segments to analyze the solar output during each week of the month. This will allow me to compare the presence of cloudy periods across the weeks.}

\texttt{\textbf{Action:} \colorbox{yellow!40}{segment\_series}, Action Input: \{Solar panel 1, \colorbox{yellow!40}{4}\}}

\texttt{...}

\texttt{\textbf{Thought:} Now that I have \colorbox{yellow!40}{segmented both time series into four weeks} and have their mean solar outputs, I can compare them to assess the presence of cloudy periods. A lower mean output during a week would indicate cloudy weather conditions.}

\texttt{...}

\texttt{\textbf{Thought:} I now know the final answer.}

\texttt{\textbf{Final Answer:} the first time series has cloudy periods for the \colorbox{yellow!40}{second} week, whereas the second time series are cloudy for the \colorbox{yellow!40}{second} and \colorbox{yellow!40}{fourth} week.}
\end{tcolorbox}



\end{document}